\documentclass[
]{ceurart}

\usepackage{listings}
\usepackage{caption}
\usepackage{subcaption}
\usepackage{soul}
\begin{document}

\copyrightyear{2023}
\copyrightclause{Copyright for this paper by its authors.
  Use permitted under Creative Commons License Attribution 4.0
  International (CC BY 4.0).}

\conference{In A. Martin, K. Hinkelmann, H.-G. Fill, A. Gerber, D. Lenat, R. Stolle, F. van Harmelen (Eds.), 
Proceedings of the AAAI 2023 Spring Symposium on Challenges Requiring the Combination of Machine Learning and Knowledge Engineering (AAAI-MAKE 2023), Hyatt Regency, San Francisco Airport, California, USA, March 27-29, 2023.}

\title{Knowledge-Based Counterfactual Queries for Visual Question Answering}

\author[1]{Theodoti Stoikou} [orcid=0000-0002-3653-0041, email=dido.stoikou@gmail.com]
\author[1]{Maria Lymperaiou}[orcid=0000-0001-9442-4186, email=marialymp@islab.ntua.gr]

\address[1]{AILS Lab,
School of Electrical and Computer Engineering,
National Technical University of Athens}

\author[1]{Giorgos Stamou}[orcid= 0000-0003-1210-9874, email=gstam@cs.ntua.gr]
\begin{abstract}
Visual Question Answering (VQA) has been a popular task that combines vision and language, with numerous relevant implementations in literature. Even though there are some attempts that approach explainability and robustness issues in VQA models, very few of them employ counterfactuals as a means of probing such challenges in a model-agnostic way. In this work, we propose a systematic method for explaining the behavior and investigating the robustness of VQA models through counterfactual perturbations. For this reason, we exploit structured knowledge bases to perform deterministic, optimal and controllable word-level replacements targeting the linguistic modality, and we then evaluate the model's response against such counterfactual inputs. Finally, we qualitatively extract local and global explanations based on counterfactual responses, which are ultimately proven insightful towards interpreting VQA model behaviors. By performing a variety of perturbation types, targeting different parts of speech of the input question, we gain insights to the reasoning of the model, through the comparison of its responses in different adversarial circumstances. Overall, we reveal possible biases in the decision-making process of the model, as well as expected and unexpected patterns, which impact its performance quantitatively and qualitatively, as indicated by our analysis. 

\end{abstract}


\maketitle

\section{Introduction}

The indisputable rise in popularity of visiolinguistic (VL) learning \cite{survey3, survey7, survey-kvl} has offered a variety of impressive model implementations to the community in a short time \cite{visualbert, vilt, vilbert, showaskattend, flava, albef}. Visual Question Answering (VQA) is a VL task that has obtained a fundamental role in the evolution of various interactive VL AI systems, such as Visual Dialogue \cite{dialog-gen}, Text-Image Retrieval \cite{retrieval} and Visual Commonsense Reasoning \cite{vcr}. 
To this end, there is an extensive range of real-world applications that benefit significantly from the new advances around the VQA task, such as aiding systems for visually impaired individuals \cite{blind, blind2} and self-driving cars \cite{cars}.

VQA involves a textual question \textit{q} from a pre-defined question set $Q$ accompanied by an image \textit{I}, the interaction of which yields a textual answer \textit{a}.
The race for continuously advancing VQA model performance unavoidably results in leaving open issues, especially attributed to the black-box nature of state-of-the-art implementations \cite{rubi, causalvqa, robustvqa, bottom-up, vqav2}. This limited access to the reasoning that such models follow to make decisions emphasizes the risk of an arbitrary behavior on their behalf. This peril lies mainly in the possibility of bias integration, decisions that lack the proper focus, as well as the absence of explainability and fairness of results. Especially when pivotal decisions are made based on this type of systems, their opacity renders them impractical, and at times hazardous, for most applications. This uncertainty indicates the need for new robustness evaluation methods, that prioritize the transparency of VQA models. 

Different approaches to debiasing and explainability of VQA models focus on diverse aspects of the issue. For example, \cite{coin} examines VQA robustness and explainability by addressing transformations on the visual modality, as they attribute the problem mostly to the visual bias as occurring from unwanted correlations between image concepts. In general, existing works primarily focus on the effect of \textit{visual bias} rather than the impact of \textit{linguistic bias}, as a reason behind the lack of robustness in VQA models. Other works follow attention-based strategies that require extensive knowledge on the model architecture, thus they cannot handle efficiently the \textit{black-box} nature of these systems. To this end, various model-specific approaches are proven to be fruitful in their strict framework, but lack the capability to be generalized to the evaluation of any other model, thus limiting their efficiency scope to just one specific case.  

We argue that resolving explainability challenges in VQA models calls for a counterfactual approach, implemented as word-level perturbations on the questions \textit{q}; thus, we diverge from the well-sought exploration of the visual modality, examining the role of the language on possible biases and spurious correlations hidden in VQA models, while tracing and interpreting their opaque decision-making process. Our proposed counterfactual perturbations are framed as: \textit{"What is the response of the VQA model if we substitute word X with word Y in question q?"}
Specifically, by viewing words as concepts, we perform the minimum possible feasible transformation to stimulate a change in the model's response; then, insightful comparisons are made by recording the model's behavior to various such transformations. The counterfactual perturbations that we perform are fully guided by the deterministic assurance of \textit{hierarchical knowledge} structures. By deploying these knowledge sources, we provide transformations that are not only optimally targeted to each specific linguistic concept, but are also fully \textit{explainable} in terms of the strategy followed for their implementation.

Starting with the observation of \textit{local model responses} in different linguistic perturbations for a single data sample, we further identify \textit{global patterns} that refer to the overall behavior of the model when faced with a specific set of perturbed concepts. Following this, we propose global rules that characterize the response of a model and can underline its weaknesses, by indicating what concepts could harm its robustness and certainty. This process reveals possible biases that a model has integrated, and hence attributes explanations as to why a particular answer is generated in place of another one; thus, we are able to obtain insights to the reasoning process of a model, without the need of access to the model's inner architecture.
Our method is generalizable to any VQA model and corresponding suitable dataset, as it approaches the issue in a totally model-agnostic strategy. To sum up, we contribute to the following:


\begin{enumerate}
    \item We design counterfactual inputs applying a variety of structured \textit{word-level replacements} on the questions $q \in Q$, as instructed by hierarchical knowledge sources. Our approach is model-agnostic, as we treat any VQA model as a black box.
    \item We obtain \textit{local explanations} derived from unexpected model responses to counterfactual \textit{q} inputs.
    \item By summarizing local model behaviors for all $q \in Q$, we extract some \textit{global explanations} that reveal the overall model response to each of the designed counterfactual inputs.
\end{enumerate}

\section{Visual Question Answering (VQA)}
\label{sec:vqa}
\begin{figure}[h!]
    \centering
    \includegraphics[scale=0.7]{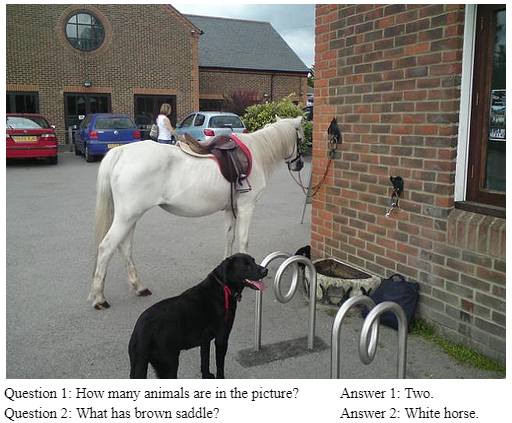}
    \caption{Example of image and free-form questions retrieved from the Visual Genome dataset \cite{visualgenome}, targeted to the VQA task. The displayed answers were given as response by the ViLT model \cite{vilt}.}
    \label{fig:horse}
\end{figure}

Visual Question Answering (VQA), first introduced in \cite{vqa}, lies in the category of multimodal learning tasks, as it receives both visual and linguistic modalities as input. Specifically, a VQA model $M$ receives images $i$ from a set \textit{I} and relevant questions $q$ belonging to a predefined question set $Q$, and is expected to accurately answer those $q$ by providing a natural language answer $a$. The aforementioned answers can be either open-ended (generated by $M$) or belong to a set of pre-defined candidates $A$. 
In general, questions $q$ have an arbitrary nature and they enclose different computer vision sub-problems, such as object recognition and detection, attribute and scene classification, as well as counting \cite{Kafle_2017}. Furthermore, more intricate questions concern more complex processes, such as spatial relationships among objects and commonsense reasoning. 
According to each specific case, visual questions $q$ selectively target different areas of an image $I$, including background details and underlying context. In accordance, the focus regarding the linguistic input lies in different word concepts depending on each image-question pair. An example of the VQA task, including an image $I$ and related questions $q$, as well as the answers $a$ that a VQA model $M$ returns to these questions, is demonstrated in Fig. \ref{fig:horse}.

Our counterfactual approach regarding linguistic substitutions revolves around the following fundamental question: \textit{"What is the response of $M$ if we substitute word X with word Y in question q?"}. The implemented counterfactual $X\rightarrow Y$ substitution should be semantically \textit{minimal} and linguistically \textit{feasible}. \textit{Minimality} refers to substitutions that maintain a meaning close to the meaning of the original word \textit{X}. For example, synonym words preserve this minimality constraint. In order to ensure semantic minimality of substitutions, we leverage lexical knowledge sources (such as WordNet \cite{wordnet}), which can provide the minimum possible $X\rightarrow Y$ transitions by selecting the closest concept $Y$ to concept $X$ that respects certain constrains. Linguistic \textit{feasibility} instructs meaningful substitutions which always involve the same part of speech (POS); for example, nouns can only be substituted by nouns but not by verbs. In total, such $X\rightarrow Y$ substitutions are applied on the whole $Q$ set, targeting one POS at a time.

Such counterfactual questions are able to trigger alternative model responses. Therefore, a $X\rightarrow Y$ concept substitution in the input may result in an alternative $X'\rightarrow Y'$ response in the output, or not. Probing potential output changes is highly informative with respect to the reasoning process followed by the model $M$, highlighting concepts or concept families that are more or less influential to the decision-making process of $M$. Hence, the counterfactual substitutions implemented on $q$ provide useful explanations for the model's observed behavior and enhance its interpretability extent, while also handling it as a black-box structure.

\section{Related work}
\paragraph{VQA models}
Since the introductory work on VQA \cite{vqa}, several endeavors have extended this paradigm, either by suggesting advanced model architectures or by proposing more challenging datasets. State-of-the-art models addressing the VQA task are mainly based on VL transformer backbones; thus, models such as ViLBERT \cite{vilbert}, VisualBERT \cite{visualbert}, FLAVA \cite{flava}, ALBEF \cite{albef},  ViLT \cite{vilt} and others
have dominated the recent VQA literature demonstrating rapid improvements on relevant benchmark datasets.
Regarding datasets, improvements on the original VQA (VQA-v2) suggest adding similar image pairs corresponding to the same question $q$, but leading to diverging answers \cite{vqav2}. Visual Genome (VG) is another large scale dataset including numerous scene images, object, attribute and relationship annotations, as well as visual question-answer pairs \cite{visualgenome}. Our approach is tested on both VQA-v2 and VG datasets.
Other popular VQA datasets are Flickr30k-Entities \cite{flickr-entities}, COCO-QA \cite{coco_qa}, Visual7W \cite{Visual7W} and others. For a detailed analysis on VQA and relevant topics we refer readers to recent specialized survey papers \cite{survey3, survey7, survey-kvl, survey-vqa, survey-vqa2, survey-vqa3}.

\paragraph{Explainability in VQA}
Regarding the research topic of explainability and robustness in Visual Question Answering \cite{leveragingdepth_explainability, multimodal_interactive_expl, robustness}, multiple efforts have been proposed, including attention maps \cite{coattention, fantastic}, and other model-specific approaches \cite{nlx}. The strategy through counterfactuals is a rather new one, while already existing attempts focus on visual perturbations \cite{coin}, masking \cite{robustvqa, robustvqatraining}, introducing counterfactuals in the training stage \cite{counterfactuallearning, robustvqatraining} and relationship-driven approaches between original and counterfactual samples \cite{contrast}.

\paragraph{Linguistic perturbations}

There is a variety of prior works that perform word-level linguistic perturbations, even though they target purely linguistic tasks, mostly text classification \cite{ren, eda, bae, textattack, aeda}, but also semantic similarity \cite{coling22} and machine translation \cite{paeg}. Our perturbations regarding synonym replacement and random noun deletion are inspired by \cite{eda}, guiding substitutions with the usage of WordNet \cite{wordnet}.
Color perturbations are adapted from \cite{coling22}, upon which we construct an appropriate hierarchy based on color distance. The rest of our implemented replacements involving noun and verb substitutions are completely novel ideas.

\section{Method}
The input of our framework consists of a dataset $D$ that contains aligned images \textit{I}, textual questions forming a set $Q$ and candidate textual answers \textit{A}. We will later present results on Visual Genome (VG) \cite{visualgenome} and VQA-v2 \cite{vqav2}, which satisfy these requirements.

We select ViLT \cite{vilt} as a proof-of-concept pre-trained VQA model $M$. Nevertheless, our proposed method is not restricted to ViLT, as it only considers inputs (questions) and outputs (answers). ViLT receives a question $q \in Q$ and an image $i \in I$ from $D$ and then
generates an answer $a$, rather than selecting one of the candidates $a \in A$; since this behavior is inherent to several VQA models, it is important to allow looser definitions of the accuracy metric. To be more precise, ViLT produces outputs $a$ in the form of natural language text, and there are many different ways of expressing the same answer in English.
In this case, heuristically comparing the generated answer with the ground truth answer from \textit{A} defines if the prediction of $M$ is accurate or not. By repeating the same prediction process for all \textit{Q, I} pairs, and by obtaining successful or unsuccessful answers for them, we finally extract an accuracy score  $\textit{acc}_Q$, reflecting the ratio of correct answers over all generated answers. 



Our word substitutions are guided from external knowledge sources, targeting different parts of speech (nouns, verbs and adjectives) at a time. Specifically, WordNet knowledge graph \cite{wordnet} provides hierarchical relationships between an abundance of common words widely present in VG and VQA-v2 vocabularies. Therefore, substitution pairs are created by connecting specific words with their WordNet matches, respecting hierarchical relationships as described in Section \ref{sec:perturb}.
Furthermore, we extend the Matplotlib color\footnote[2]{\href{https://matplotlib.org/3.3.3/gallery/color/named_colors.html}{Matplotlib colors}} relationships presented in \cite{coling22}, forming a hierarchy of \textit{color relatedness}. This color hierarchy is based on color distances according to the RGB value of each Matplotlib color, with more details provided in Section \ref{sec:perturb}.7

\begin{figure}[ht!]
\vspace{-0.7cm}
    \centering
    \includegraphics[width=\textwidth]{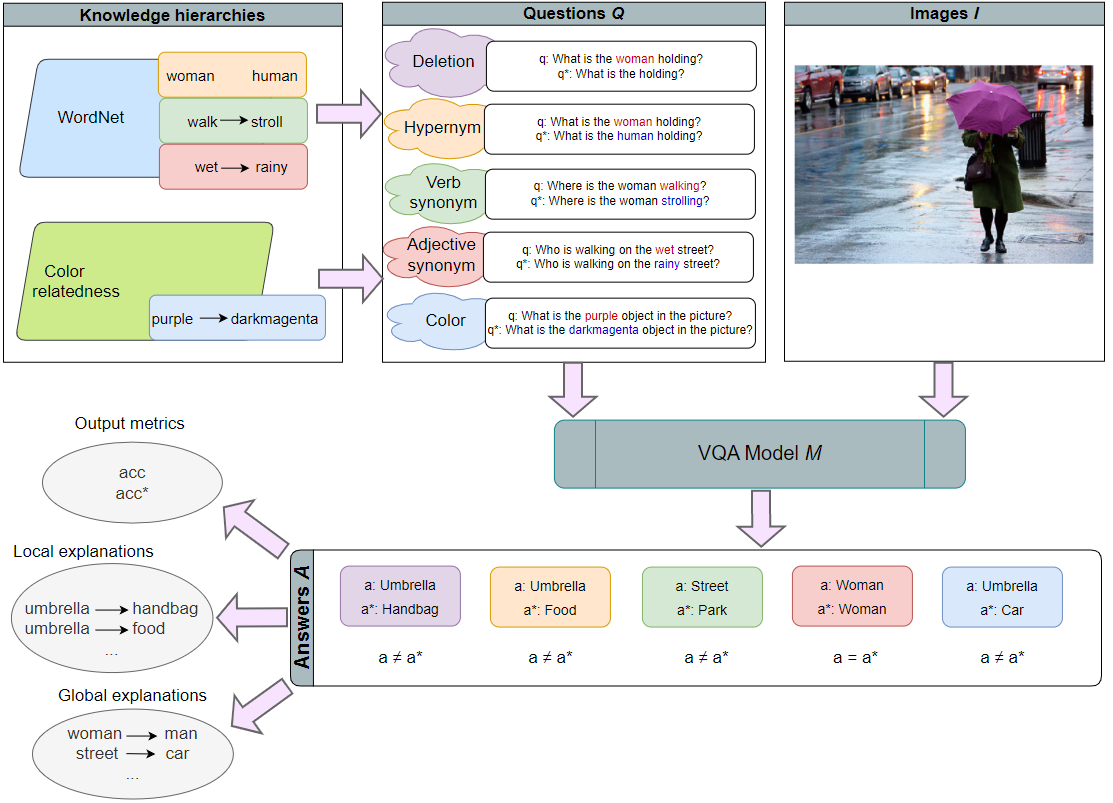}
    \caption{Overview of our proposed knowledge-based counterfactual VQA framework.}
    \label{fig:overview}
\end{figure}

We then proceed with applying our designed perturbations on dataset questions $q \in Q$, resulting in \textit{counterfactual questions} $q^* \in Q^*$.  
For each substitution, we obtain the \textit{counterfactual question accuracy} $\textit{acc}^*_Q$, as the response of  $M$ to the counterfactual questions $q^* \in Q^*$, and we compare it to the ground truth accuracy scores $\textit{acc}_Q$. 
Throughout this process, we evaluate whether and how the response of $M$ changes by measuring the difference between $\textit{acc}_Q$ and $\textit{acc}^*_Q$, as an indicator of its robustness against replacements with semantically related concepts. 

Even though useful for benchmarking reasons, standalone accuracy scores are not informative enough to explain \textit{why} we observe such differences between original $q$ and counterfactual inputs $q^*$. To this end, we separately examine samples where the generated $a$ changes under the presence of a perturbation, obtaining local explanation in the form \textit{if concept changes in q, then a -erroneously- changes}. The aggregation of such local rules leads to \textit{global explanations}, deriving \textit{if-then} relationships that apply to multiple samples of $D$. 

We present a visual outline of our approach in Figure \ref{fig:overview}. Words from $q \in Q$ colored in \textcolor{red}{red} denote the concepts to be substituted, while words in \textcolor{blue}{blue} indicate the knowledge-driven substitutions that lead to counterfactual questions $q^* \in Q^*$.


\subsection{Perturbations}
\label{sec:perturb}
In our work, we perform a variety of substitutions or deletions in the linguistic representation of \textit{Q}. This counterfactual strategy exploits multiple and diverse morphological attributes of \textit{Q} and attempts to demonstrate the semantics that most affect the model's response $a$. Our target is to stimulate an altered response of $M$ through these counterfactual perturbations, in order to identify how the behavior of $M$ changes when faced with different concepts. Thus, we can infer potential biases or points of weak robustness in $M$. The aforementioned substitutions can be divided into the following categories, based on the knowledge source used and the targeted part of speech. A summary and representative examples of substitutions are provided in Table \ref{tab:examples}.

\textbf{A. Wordnet hierarchy:} The knowledge-driven word substitutions involve replacing a noun word from questions $q \in Q$ with a hierarchically related word (\textit{hyponym}, \textit{hypernym}, \textit{sibling}), or verbs and adjectives with their \textit{synonyms}. The quality and relevance of our substitutions are reassured by the use of the deterministic structure of the Wordnet hierarchy, guaranteeing controllable and optimal word-level substitutions. 
\begin{itemize}
 \item \textbf{Synonyms:} We employ synonym transformations on adjectives and verbs of the original questions $q \in Q$. For example, "talk" and "speak" are \textit{synonym verbs} according to WordNet, while "small" and "minuscule" are \textit{adjective synonyms}. 
 \item \textbf{Hypernyms - Hyponyms:} More \textit{general}, as well as more \textit{specific} noun concepts are provided via WordNet in the form of \textit{hypernyms} and \textit{hyponyms} respectively.
 For example, a given noun word (e.g. "dog") we can extract its immediate noun hypernyms (e.g. "canine"), or its immediate hyponyms (e.g. "labrador"). 
\item \textbf{Siblings:} We construct noun \textit{sibling} substitutions by traversing the Wordnet knowledge tree one step upwards and then one step downwards. Siblings are defined as noun entities that share the same immediate parent. For example, "carrot" and "radish" are siblings, because they both have "plant root" as their parent concept according to WordNet.
\end{itemize}

\textbf{B. Color relatedness hierarchy:}
Colors that are semantically similar, therefore presenting RGB values close to each other, will be also close within the color relatedness hierarchy. For example, "violet" and "orchid" Matplotlib colors lie close within the color hierarchy (their in-between color distance is 6.16), while "violet" and "deepskyblue" are placed far away from each other (their color distance is 207.88). Colors can be replaced with either distant or else similar colors from this color relatedness hierarchy, leading to the following \textbf{Color Maximal} and \textbf{Color Minimal} color substitutions. Both Maximal/Minimal substitutions may either involve \textit{common} colors, which already exist in the dataset or else \textit{uncommon} colors, which belong to the Matplotlib color list but not in VG/VQA-v2 vocabularies:
\begin{itemize}
    \item \textbf{Color Maximal:} On questions that mention some specific color we contradict the output $a$ of $M$ based on the input of the original question $q \in Q$ vs the output $a^*$ of the perturbed question $q^* \in Q^*$. In $q^*$ the original color is substituted with one that is greatly distant to it, such as "violet" $\rightarrow$ "deepskyblue". 
    In this category, we also challenge the model using less frequent color instances (i.e. "azure", "turquoise", "salmon"). This substitution diverges from the initial counterfactual question requesting \textit{minimal} changes; nevertheless, the comparison with related \textit{minimal} changes will highlight the differences that varying color distances impose on the final $acc^*_Q$.
    \item \textbf{Color Minimal:} In accordance with the above, we perform color substitutions with the least distant colors, such as  "violet" $\rightarrow$ "orchid".  Again we also challenge the model with less frequent color substitutions.
\end{itemize}

\textbf{C. Deletions:} We randomly select a noun in each question $q \in Q$ and remove it. 

\begin{table}[h!]
  \centering
  \caption{Question perturbations examples towards counterfactual queries.}
  \label{tab:examples}
  \begin{tabular}{c|c}
    \toprule
    Perturbation & Question \\
    \midrule
    \textbf{Original} & Do you see the white small dog? \\
    \textbf{Color Maximal} & Do you see the \textbf{black} small dog ?\\
    \textbf{Color Minimal} & Do you see the \textbf{beige} small dog ?\\
    \textbf{Synonym Adjectives} & Do you see the white \textbf{tiny} dog ?\\
    \textbf{Synonym Verbs} & Do you \textbf{watch} the white small dog ?\\
    \textbf{Hypernym Noun} & Do you see the white small \textbf{canine} ?\\
    \textbf{Hyponym Noun} & Do you see the white small \textbf{labrador} ?\\
    \textbf{Sibling Noun} & Do you see the white small \textbf{wolf} ?\\
    \textbf{Deletion Noun} & Do you see the white small \_ ?\\    
    
    \bottomrule
  \end{tabular}
\end{table}

Substitutions and deletions are an excellent way to quantify whether a VQA model $M$ understands a specific question-image pair, or if its output is greatly dependent on biased estimations. This way, we can reveal spurious correlations that are mistakenly integrated into $M$. Color substitutions are motivated by the quantity of color-related questions that exist in our input datasets (VG and VQA-v2). By interrogating $M$ with color perturbations that are greatly distant to the original color (\textbf{Color Maximal} substitutions experiment), we aim to detect whether $M$ will correctly and reasonably perceive this semantically massive change. We would expect $M$ to change its response $a^*$ in most cases of \textbf{Color Maximal} experiment; the opposite would indicate an underlying pattern of ignoring color attributes. Similarly, we perform the \textbf{Color Minimal} substitution experiment in order to investigate the model's behavior when faced with minor alterations in the color concept. We expect the substitutions from this experiment to have little to no influence on the model's response $a^*$. An opposite behavior would reveal an existing bias regarding specific colors, which would lead to the conclusion that $M$ cannot properly and robustly adapt to minor color changes and generalize accordingly. Of course, \textit{uncommon} color substitutions in both Minimal/Maximal cases impose a more difficult problem, as $M$ needs to adaptively respond to out-of-dataset color concepts.

In relation to the \textbf{Synonym} substitutions, we aim to investigate the model's ability to efficiently handle mild morphological language alterations that maintain the same meaning. In this case, failing to properly respond (providing an alternative $a^*=a$) would disclose overfitting to specific semantics, which renders the model lexically inflexible and thus non-robust to semantically negligible perturbations. The \textbf{Hypernyms}-\textbf{Hyponyms} perturbations are dedicated to depicting the model's ability to generalize and specify correspondingly, while retaining a reliable level of robustness. Hypernym and hyponym relationships are notions profoundly understood in the real world and consequently embedded in large scale datasets, which are widely used for VQA models pre-training. Thus, $M$ should also be able to properly comprehend and reason over them. Ideally, we would expect $M$ to maintain the same response for hypernyms substitutions, whereas justifiably respond in specific ways for the hyponyms replacements, taking into account the specification of meaning. The commensurate amount of specifying skill is sought to be established through \textbf{Sibling} substitution experiment. Depending on each particular case, we expect $M$ to modify or maintain its response appropriately, to confirm the level of understanding and distinguishment of different, but still related, meanings. 

Finally, we implemented the \textbf{Deletion} experiment expecting ideally the performance of $M$ to degrade. The amount of the $acc^*_Q$ decline depends on the importance of the deleted noun for the meaning of the question. Consequently, an unbiased $M$ should be able to determine this importance and act accordingly, without reaching unwarranted conclusions that are expressed through an indefensible response.

\section{Experiments}

We present results using the accuracy metric, which illustrates the extent of similarity of the model’s predicted answer to the ground truth answer, both for the original $Q$ of each dataset, as well as for the counterfactual question set $Q^*$. In our analysis, accuracy is not profound enough to provide specific situational explanations and insights on the model’s behavior, when faced with particular concepts. However, accuracy still showcases a high-level approach on the model’s efficiency fluctuations under the implemented counterfactual perturbations. Since ViLT model is trained and optimized on the VQA-v2 dataset, it is somehow expected to perform better on it compared to VG (both datasets contain similar vocabularies). This observation is indeed validated by our results presented in Tables \ref{color} \& \ref{perts}, which demonstrate a consistently higher $acc_Q$ on the former versus the latter dataset, concerning all implemented experiments. We denote that $acc_Q$ scores for each experiment contain the corresponding questions only, e.g. color experiments only contain questions that mention colors. This contributes to the differences in original $acc_Q$ scores for each experiment. 
Nevertheless, in both datasets, we notice an analogous difference between $acc_Q$ and $acc^*_Q$ per experiment  when $M$ is presented with counterfactual questions $q^* \in Q^*$. This could generally indicate the existence of underlying biases: $M$ presents a type of overfitting to the original $q \in Q$, which renders it less efficient when asked to handle minimally perturbed counterfactual questions. In all experiments, the accuracy reduction from the original $acc_Q$ to $acc^*_Q$ is approximately 15-20\% or more.

An extended depiction of the retrieved accuracies for both datasets is presented in Table \ref{color} (color-based substitutions) and Table \ref{perts} (WordNet-based substitutions and noun deletions). Specifically, for \textbf{Color Maximal} in VQA-v2 we observe a decline of 34.2\% for \textit{common} colors and a decline of 37.4\% for \textit{uncommon} ones. Even for semantically minimal substitutions, (\textbf{Color Minimal} experiment), the decline is 31\% for both \textit{common} and \textit{uncommon} colors. As for VG, we observe a decline of 38.2\% for \textit{common} colors and a decline of 52.2\% for \textit{uncommon} colors when \textbf{Color Maximal} substitutions are performed. Correspondingly, a decline of 35\% for both \textit{common} and \textit{uncommon} colors is reported for \textbf{Color Minimal} substitutions.

\begin{table*}[h!]
  \caption{Accuracies for color perturbations on VQA-v2 and Visual Genome (VG). Common refers to substitutions with in-dataset colors, while uncommon refers to substitutions involving any Matplotlib color.}
  \label{color}
  \begin{tabular}{c|cc|cc|cc}
  \toprule
   &  \multicolumn{2}{c}{$acc_Q$\%}&  \multicolumn{2}{c}{$acc^*_Q$\% (common)} &  \multicolumn{2}{c}{$acc^*_Q$\% (uncommon)} \\
    \toprule
     Perturbation & VQA-v2 & VG & VQA-v2 & VG & VQA-v2 & VG \\
    \midrule
    \textbf{Color Maximal} & 69.6 & 46.9 & 45.8 & 29.0 & 43.6 & 22.4 \\
    \textbf{Color Minimal} & 70.0 & 47.5 & 48.3 & 30.9 & 48.3 & 30.9 \\
    \bottomrule
\end{tabular}
\end{table*}

\begin{table*}[h!]
\caption{Accuracies for WordNet-based perturbations 
 on VQA-v2 and Visual Genome (VG).}
\label{perts}
\begin{tabular}{c|cc|cc|cc}
\toprule
&  \multicolumn{2}{c}{$acc_Q$\% }&  \multicolumn{2}{c}{$acc^*_Q$\% } & \multicolumn{2}{c}{$acc_Q$ reduction \%} \\
\midrule
Perturbation & VQA-v2 & VG & VQA-v2 & VG & VQA-v2 & VG\\
\midrule
\textbf{Synonym Adjectives}  & 75.1 & 47.0 & 56.9 & 37.4 & 20.6 & 20.4 \\
\textbf{Synonym Verbs}  & 76.8 & 52.3 & 64.1 & 44.4 & 16.5 & 15.1 \\
\textbf{Hypernym Noun} & 75.2 & 54.6 & 60.8 & 41.5 & 19.1 & 24.0 \\
\textbf{Hyponym Noun}  & 75.1 & 53.5 & 56.3 & 36.1 & 25.0 & 32.5 \\
\textbf{Sibling Noun}  & 76.9 & 54.0 & 54.0 & 33.4 & 29.8 & 38.1 \\
\midrule
\textbf{Deletion Noun} & 76.9 & 53.9 & 59.1 & 36.5 & 23.1 & 32.3 \\
\bottomrule
\end{tabular}
\end{table*}

The discovery of global patterns provides a more profound and targeted view on robustness of the model $M$. To this end, we note that in all the cases we studied, we are not interested in the ground truth answer of a question $q$, but rather in the differentiation between the answer $a^*$ to the counterfactual question in relation to the original answer $a$ that $M$ predicts, either if $a$ is correct or not. We select this approach since we are interested in discovering the model's change in decision-making under the presence of counterfactual inputs, which is more informative than measuring how much $a^*$ semantically deviates from the ground truth response.

Based on the thorough investigation of our experiments’ results and the aggregation of the following \textit{local explanations}, as presented in the upcoming Figures, we have deduced some meaningful \textit{global rules} that both embody the robustness of $M$ to our counterfactual questions $q^* \in Q^*$, while providing reliable explanations that reveal the model’s reasoning behind its decision-making. Furthermore, we analyze underlying existing biases of $M$ that logically derive from these global rules. 
In the following Figures, we highlight the original $q, a$ with \textcolor{red}{red} and the counterfactual $q^*, a^*$ with \textcolor{blue}{blue}.

\subsection{Color Maximal explanations}

\begin{figure}[h!]
     \centering
     \begin{subfigure}[b]{0.49\textwidth}
         \centering
         \includegraphics[width=0.9\textwidth, height=4cm]{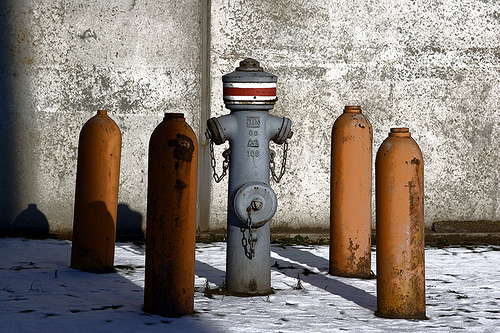}        
         \caption{\small{\textbf{q}: What is surrounding the \textcolor{red}{silver}/\textcolor{blue}{black}/\underline{\textcolor{blue}{navy}} fire hydrant? \\ \textbf{a}: \textcolor{red}{posts}/\textcolor{blue}{posts}/\underline{\textcolor{blue}{posts}}.}}
         \label{11268}
     \end{subfigure}
     \hfill
     \begin{subfigure}[b]{0.49\textwidth}
         \centering
         \includegraphics[width=0.9\textwidth, height=4cm]{images/1765.png}
         \caption{\small{\textbf{q}: How many glasses have \textcolor{red}{red}/\textcolor{blue}{gold}/\underline{\textcolor{blue}{darkturquoise}} wine? \\ \textbf{a}: \textcolor{red}{6}/\textcolor{blue}{0}/\underline{\textcolor{blue}{0}}.}}
         \label{1765}
     \end{subfigure}
        \caption{Local explanations for \textbf{Color Maximal} counterfactual perturbations.}
        \label{Maximal photos}
\end{figure}

In \textbf{Color Maximal} substitutions, we notice that $M$ erroneously maintains the same answer $a^*=a$ when we replace the colors \textit{gray} and \textit{silver} with any other semantically maximal color, either common or uncommon (underlined). Therefore, we detect a bias in the model related to these two colors, as it does not make logical decisions after replacing them with others and does not properly reason over this substitution. A relevant example is presented in Figure \ref{11268}.

However, contrary to the above, $M$ logically revises its answers when we replace the colors \textit{green} and \textit{red} with any distant colors, either common or uncommon (underlined) as presented in Figure \ref{1765}. Therefore, the model recognizes and qualitatively understands these substitutions and has not incorporated any problematic attachment regarding these two colors.

This observation denotes that $M$ is more sensitive towards intense and visually distinct colors and rather bypasses changes involving more neutral ones, focusing on object identities (e.g. "fire hydrant" and "posts" of Figure \ref{11268}). A more uncertain $a^*$ (e.g. the model's answer $a^*$ could be "nothing") would be more suitable, if all question semantics were equally taken into account.

\subsection{Color Minimal explanations}

\begin{figure}[h!]
     \centering
     \begin{subfigure}[b]{0.49\textwidth}
         \centering
         \includegraphics[width=0.9\textwidth, height=4.3cm]{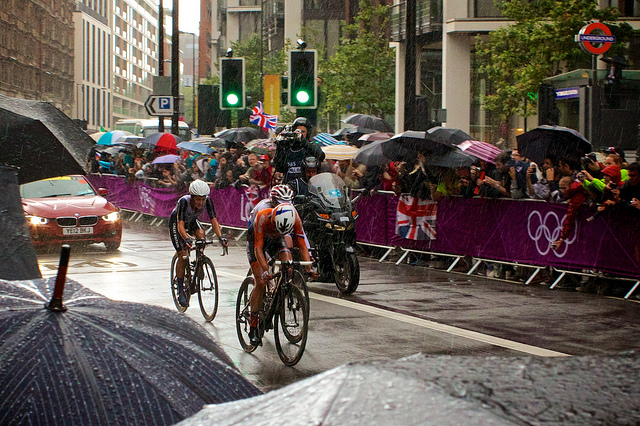}        
         \caption{\small{\textbf{q}: What organization's logo is on the \textcolor{red}{purple}/\textcolor{blue}{blue}/\underline{\textcolor{blue}{plum}} banner? \\ \textbf{a}: \textcolor{red}{olympics}/\textcolor{blue}{olympics}/\underline{\textcolor{blue}{olympics}}.}}
         \label{6536}
     \end{subfigure}
     \hfill
     \begin{subfigure}[b]{0.49\textwidth}
         \centering
         \includegraphics[width=0.9\textwidth, height=4.3cm]{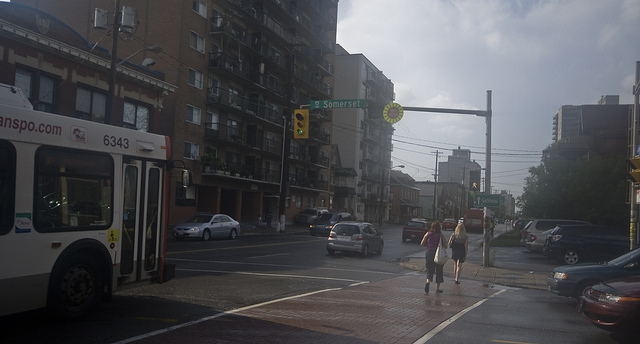}
         \caption{\small{\textbf{q}: What is being held \textcolor{red}{green}/\textcolor{blue}{forestgreen}/\underline{\textcolor{blue}{olive}}? \\ \textbf{a}: \textcolor{red}{light}/\textcolor{blue}{bus}/\underline{\textcolor{blue}{bag}}.\vspace{\baselineskip}}}
         \label{388}
     \end{subfigure}
        \caption{Local explanations for \textbf{Color Minimal} counterfactual perturbations.}
        \label{Minaml photos}
\end{figure}

Based on our experiments on semantically minimal color substitutions, we derive the following global rule: When we replace the colors \textit{gray} and \textit{purple} with any other closely related color, $M$ tends to give the same answer $a^*=a$. Therefore, $M$ does not give due importance to this change of colors, a fact that highlights a robust behavior related to the two aforementioned colors. The model maintains this invariant behavior equally when we perform replacements with \textit{common} colors or with \textit{uncommon} ones, as presented in Figure \ref{6536}.

In contrast, $M$ redefines its answers when we replace the \textit{green} color with any other, \textit{common} or \textit{uncommon}, semantically similar color. Consequently, it is being confused by such minimal changes, failing to provide a meaningful answer, as shown in Figure \ref{388}. Even a more uncertain answer (e.g. "nothing") to counterfactual questions would be more suitable compared to the semantically divergent ones returned ("bus" and "bag" instead of "light").
Likewise, $M$ presents a similar change in behavior when we replace the \textit{pink} color with common minimal colors and the \textit{silver} color with uncommon ones.

\subsection{Synonym Adjectives explanations}
\begin{figure}[h!]
     \centering
     \begin{subfigure}[b]{0.2\textwidth}
         \centering
         \includegraphics[width=\textwidth, height=2.6cm]{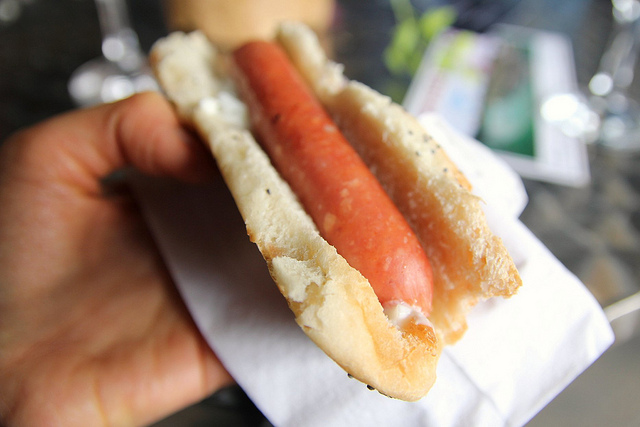}
         \caption{\small{\textbf{q}: Is this a \textcolor{red}{hot}/\textcolor{blue}{raging} dog? \\ \textbf{a}: \textcolor{red}{yes}/\textcolor{blue}{no}.\vspace{\baselineskip}\vspace{\baselineskip}}}
         \label{138}
     \end{subfigure}
     \hfill
     \begin{subfigure}[b]{0.2\textwidth}
         \centering
         \includegraphics[width=\textwidth, height=2.6cm]{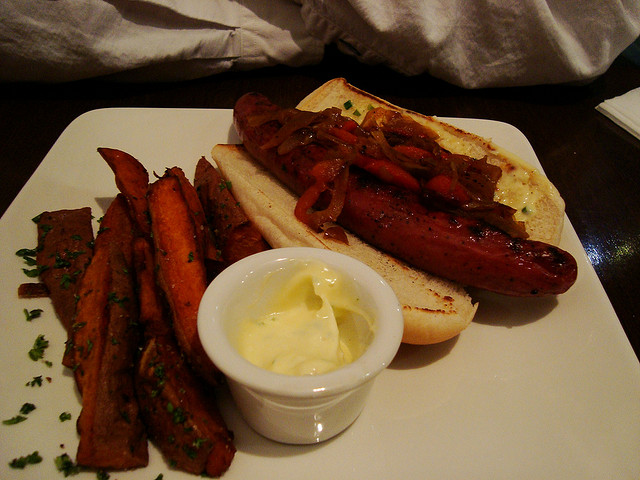}        
         \caption{\small{\textbf{q}: Of what meal is this kind of food \textcolor{red}{typical}/\textcolor{blue}{distinctive}? \\ \textbf{a}: \textcolor{red}{lunch}/\textcolor{blue}{hot dog}.}}
         \label{274}
     \end{subfigure}
     \hfill
     \begin{subfigure}[b]{0.2\textwidth}
         \centering
         \includegraphics[width=\textwidth, height=2.6cm]{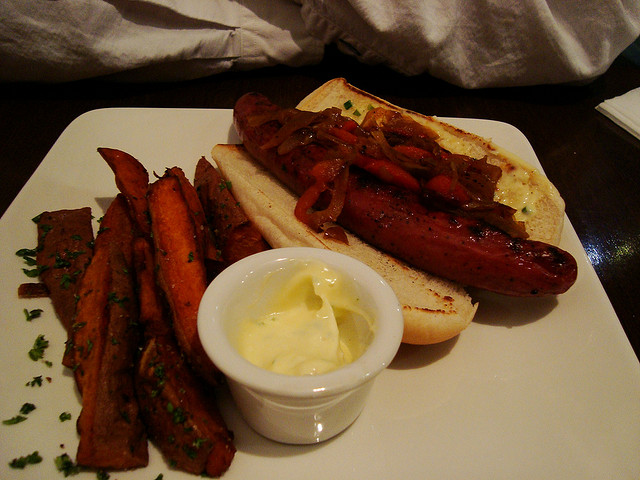}
         \caption{\small{\textbf{q}: How \textcolor{red}{delicious}/\textcolor{blue}{delightful} does this look? \\ \textbf{a}: \textcolor{red}{very}/\textcolor{blue}{not very}.\vspace{\baselineskip}}}
         \label{285}
     \end{subfigure}
     \hfill
     \begin{subfigure}[b]{0.2\textwidth}
         \centering
         \includegraphics[width=\textwidth, height=2.6cm]{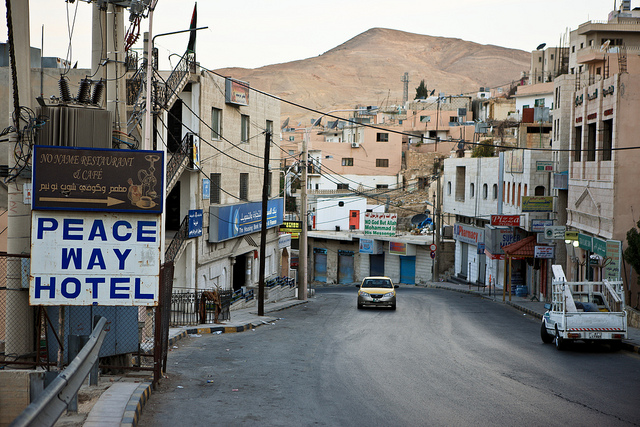}
         \caption{\small{\textbf{q}: Is this a \textcolor{red}{small}/\textcolor{blue}{little} town? \\ \textbf{a}: \textcolor{red}{yes}/\textcolor{blue}{yes}.\vspace{\baselineskip}\vspace{\baselineskip}}}
         \label{347}
     \end{subfigure}
     \hfill
     \begin{subfigure}[b]{0.15\textwidth}
         \centering
         \includegraphics[width=\textwidth, height=2.6cm]{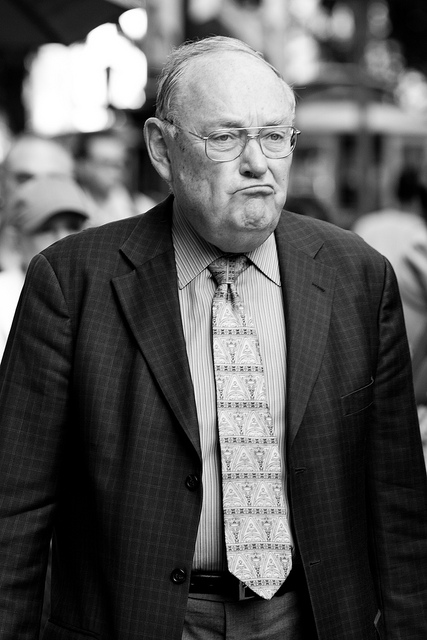}
         \caption{\small{\textbf{q}: Does the man look \textcolor{red}{happy}/\textcolor{blue}{felicitous}? \\ \textbf{a}: \textcolor{red}{no}/\textcolor{blue}{yes}.\vspace{\baselineskip}}}
         \label{20}
     \end{subfigure}
        \caption{Local explanations for \textbf{Synonym Adjectives} counterfactual perturbations.}
        \label{Adj photos}
\end{figure}

In general, adjective-noun pairs present in questions contain some joint special conceptual meaning which differs from the independent meaning of adjectives when they exist autonomously and separately in a sentence. In this case, we notice that the model $M$ varies its answer $a^*$ when we implement a synonym substitution of its question adjectives. This finding suggests that $M$ can qualitatively perceive the meaning of such adjective-noun pairs and differentiate its response accordingly, as presented in Figure \ref{138}.

Another finding is related to the ability of $M$ to correctly adjust its answer, when it is presented with a lexically correct synonym to an adjective, which, however, is not quite appropriate for the given linguistic environment of the question. Accordingly, we conclude that $M$ is capable of understanding the meaning of an adjective in relation to the context of the sentence ("typical food" is meaningful, but "distinctive food" is not), as presented in Figure \ref{274}.

In addition, we derive a global rule that concerns the behavior of $M$ when we replace an adjective having multiple meanings with one of its synonyms, which, although it is optimal with respect to the aforementioned meanings, is however not suitable to the semantic context of the substituted adjectives (such as "delicious" vs "delightful"). We note that in this case, $M$ demonstrates a stable behavior against such substitutions, which proves that it is able to reason over adjectives in a contextualized manner, without being fooled by synonyms not suitable to the exact context of the question $q$.
An example of this observation is provided in Figure \ref{285}.

Size-related adjective substitutions are demonstrated in Figure \ref{347}. We observe that $M$ is particularly robust to such substitutions, therefore correctly capturing the underlying meaning without being biased towards specific words. This global rule demonstrates the flexibility of $M$ towards appropriately handling semantically and contextually equivalent adjective substitutions.

Finally, $M$ is proven to be unstable when it has to handle rare or difficult synonyms of adjectives, as the ones shown in Figure \ref{20}. Consequently, it presents a lexical weakness in handling such rare adjectives and possibly a bias in specific words that are more familiar to it.

\subsection{Synonym Verbs explanations}

\begin{figure}[h!]
     \centering
     \begin{subfigure}[b]{0.2\textwidth}
         \centering
         \includegraphics[width=\textwidth, height=2.6cm]{images/46.png}
         \caption{\small{\textbf{q}: What \textcolor{red}{says}/\textcolor{blue}{state} STAPLES? \\ \textbf{a}: \textcolor{red}{nothing}/\textcolor{blue}{New York}.}}
         \label{46}
     \end{subfigure}
     \hfill
     \begin{subfigure}[b]{0.2\textwidth}
         \centering
         \includegraphics[width=\textwidth, height=2.6cm]{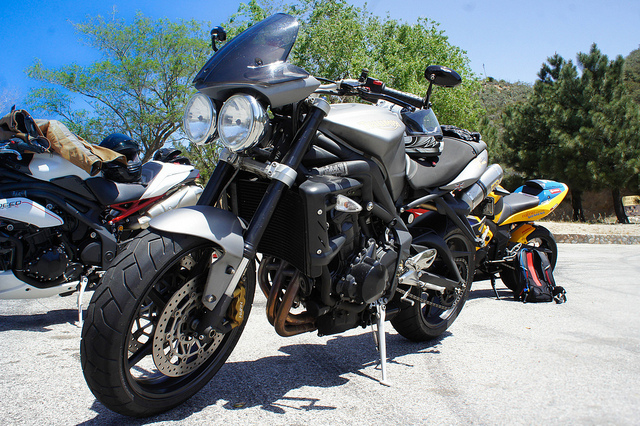}        
         \caption{\small{\textbf{q}: Do you \textcolor{red}{see}/\textcolor{blue}{understand} any motorcycle helmets? \\ \textbf{a}: \textcolor{red}{no}/\textcolor{blue}{yes}.}}
         \label{527}
     \end{subfigure}
     \hfill
     \begin{subfigure}[b]{0.2\textwidth}
         \centering
         \includegraphics[width=\textwidth, height=2.6cm]{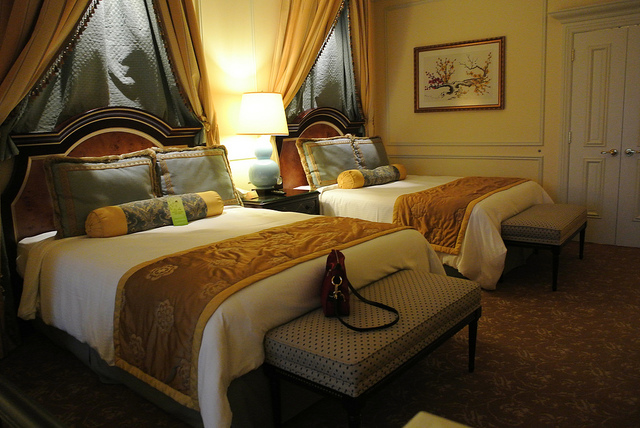}
         \caption{\small{\textbf{q}: Are the walls \textcolor{red}{done}/\textcolor{blue}{made} in a summery color? \\ \textbf{a}: \textcolor{red}{yes}/\textcolor{blue}{yes}.\vspace{\baselineskip}}}
         \label{37}
     \end{subfigure}
     \hfill
     \begin{subfigure}[b]{0.2\textwidth}
         \centering
         \includegraphics[width=\textwidth, height=2.6cm]{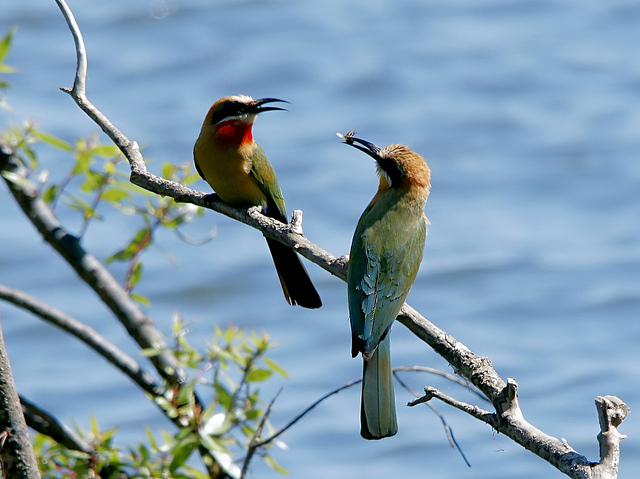}
         \caption{\small{\textbf{q}: What kind of birds are \textcolor{red}{pictured}/\textcolor{blue}{visualized}? \\ \textbf{a}: \textcolor{red}{parrots}/\textcolor{blue}{parrots}.}}
         \label{2419}
     \end{subfigure}
        \caption{Local explanations for \textbf{Synonym Verbs} counterfactual perturbations.}
        \label{Verbs photos}
\end{figure}

Regarding substitutions involving verb synonyms, $M$ is not particularly stable when dealing with substitutions of verbs that present multiple meanings, as presented in Figure \ref{46}. This indicates a difficulty in distinguishing the correct and desired meaning among multiple ones.

An even more specific rule we extract is that $M$ falsely changes its original answer $a$ when we replace the verb "see" with a qualitative synonym of it. A relevant example is provided in Figure \ref{527}. This finding indicates an unwanted attachment of the model to the word "see", which is interpreted as bias. This is an example of a more general situation, where optimal synonyms may not be the best choice for a synonym in a specific contextual setting. In these kinds of instances, the model tends to change its response, as observed in this particular case.

The model $M$ presents satisfactory robustness when it has to deal with easy or common verbs of the English vocabulary, which means that it has acquired a certain degree of versatility in simple vocabulary challenges, as shown in Figure \ref{37}.

Finally, $M$ is rather stable when the replaced verb
corresponds to a noun counterpart (e.g. picture -verb-, picture -noun-) or even adjective counterpart (e.g. pictured), such as the ones of Figure \ref{2419}. Consequently, $M$ is capable of capturing the general sense of such verbs in the context of the question; equivalent substitution of corresponding nouns (picture$\rightarrow$visualization) or adjectives (pictured$\rightarrow$visualized) would mostly yield the same counterfactual response $a^*$.

\subsection{Hypernym Noun explanations}

\begin{figure}[h!]
     \centering
     \begin{subfigure}[b]{0.3\textwidth}
         \centering
         \includegraphics[width=\textwidth, height=3.3cm]{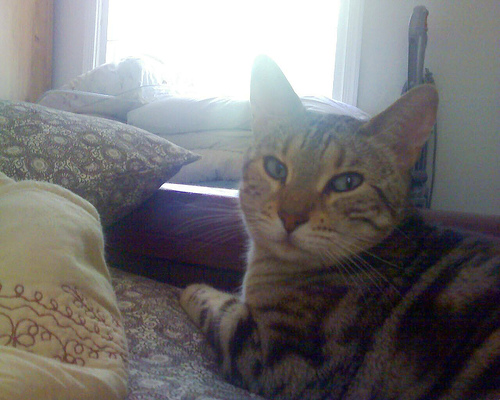}
         \caption{\small{\textbf{q}: Where is the \textcolor{red}{cat}/\textcolor{blue}{feline}? \\ \textbf{a}: \textcolor{red}{bed}/\textcolor{blue}{bed}.\vspace{\baselineskip}}}
         \label{713}
     \end{subfigure}
     \hfill
     \begin{subfigure}[b]{0.3\textwidth}
         \centering
         \includegraphics[width=\textwidth, height=3.3cm]{images/784.png}        
         \caption{\small{\textbf{q}: Are all the players wearing black \textcolor{red}{shirts}/\textcolor{blue}{garment}? \\ \textbf{a}: \textcolor{red}{no}/\textcolor{blue}{yes}.}}
         \label{784}
     \end{subfigure}
     \hfill
     \begin{subfigure}[b]{0.3\textwidth}
         \centering
         \includegraphics[width=\textwidth, height=3.3cm]{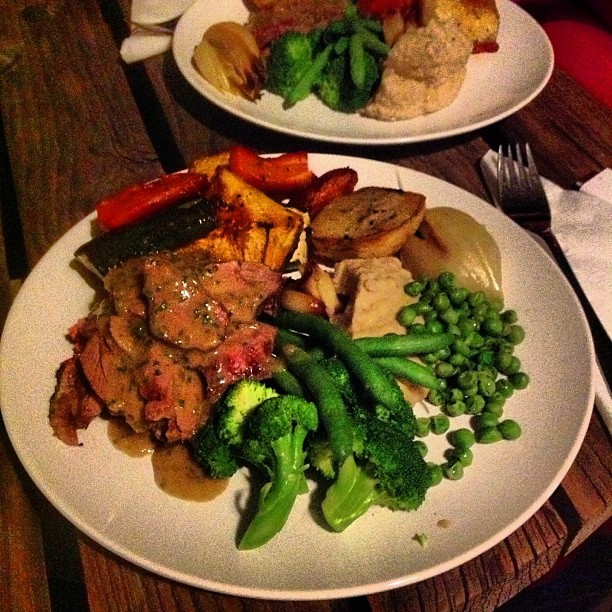}
         \caption{\small{\textbf{q}: Are there multiple vegetables on the \textcolor{red}{plate}/\textcolor{blue}{base}? \\ \textbf{a}: \textcolor{red}{yes}/\textcolor{blue}{no}.}}
         \label{863}
     \end{subfigure}
        \caption{Local explanations for \textbf{Hypernym Noun} counterfactual perturbations.}
        \label{Verbs photos}
\end{figure}

Throughout our Hypernym Noun substitutions, we conclude that $M$ is particularly robust against substitutions involving living creatures, such as animals or humans, which shows that it can properly reason over hierarchical relationships governing such concepts, as in Figure \ref{713}.

On the contrary, $M$ does not clearly distinguish between concepts related to types of clothing, i.e. it tends to erroneously change its answer when replaced with a broader concept. Consequently, $M$ does not generalize well on such entities and a bias towards more specific and clear types of clothing emerges. A relevant example is presented in Figure \ref{784}.

As an extension of the above, $M$ exhibits instability in hypernym substitutions that are very broad, inclusive, and polysemous. Therefore, when we replace a noun with an optimal hypernym that presents much greater conceptual generality, $M$ is unable to qualitatively perceive the hierarchical relation that governs them, outputting a wrong answer, as in Figure \ref{863}.

\subsection{Hyponyms Noun explanations}

\begin{figure}[h!]
     \centering
     \begin{subfigure}[b]{0.3\textwidth}
         \centering
         \includegraphics[width=\textwidth, height=3cm]{images/85.png}
         \caption{\small{\textbf{q}: Are the \textcolor{red}{animals}/\textcolor{blue}{acrodont} eating? \\ \textbf{a}: \textcolor{red}{yes}/\textcolor{blue}{yes}.}}
         \label{85}
     \end{subfigure}
     \hfill
     \begin{subfigure}[b]{0.3\textwidth}
         \centering
         \includegraphics[width=\textwidth, height=3cm]{images/784.png}        
         \caption{\small{\textbf{q}: Are all the players wearing black \textcolor{red}{shirts}/\textcolor{blue}{camise}? \\ \textbf{a}: \textcolor{red}{no}/\textcolor{blue}{yes}.}}
         \label{784b}
     \end{subfigure}     
     \hfill
     \begin{subfigure}[b]{0.3\textwidth}
         \centering
         \includegraphics[width=\textwidth, height=3cm]{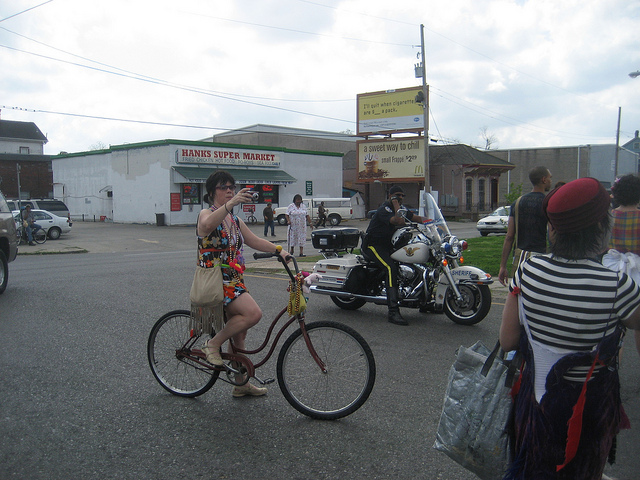}
         \caption{\small{\textbf{q}: What are objects behind the \textcolor{red}{motorcycles}/\textcolor{blue}{minibike}? \\ \textbf{a}: \textcolor{red}{sign}/\textcolor{blue}{sign}.}}
         \label{2820}
     \end{subfigure}         
        \caption{Local explanations for \textbf{Hyponyms Noun} counterfactual perturbations.}
        \label{Verbs photos}
\end{figure}

Similar to the hypernyms substitution experiment, $M$ is able to appropriately respond in cases where the substitutions of hyponyms refer to living entities. Therefore, the specialization in more specific living entities concepts is properly perceived, as presented in Figure \ref{85}.

Correspondingly, $M$ also shows stability in the substitutions of hyponyms that represent articles of clothing. Therefore, it specializes skillfully in more specific cloth-related entities and according to the case, it appropriately changes its response by adapting to the change. A relevant example is presented in Figure \ref{784b}.

On the contrary, the model does not demonstrate robustness to hyponym substitutions referring to means of transport. Consequently, the model is biased toward such broader concepts and fails to adequately understand their specialization, as shown in Figure \ref{2820}.

\subsection{Sibling Noun explanations}

\begin{figure}[h!]
     \centering
     \begin{subfigure}[b]{0.19\textwidth}
         \centering
         \includegraphics[width=\textwidth, height=2.5cm]{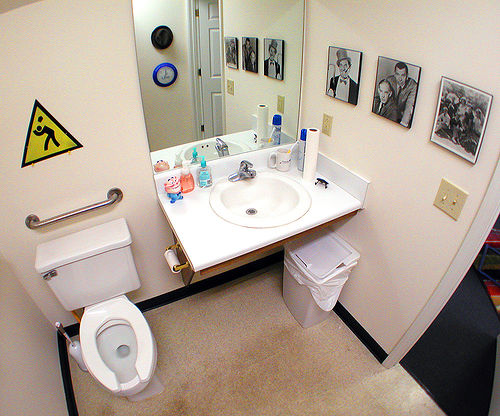}        
         \caption{\small{\textbf{q}: Is the \textcolor{red}{bathroom}/\textcolor{blue}{workroom} organized? \\ \textbf{a}: \textcolor{red}{yes}/\textcolor{blue}{yes}.\vspace{\baselineskip}}}
         \label{2605}
     \end{subfigure}
     \hfill
     \begin{subfigure}[b]{0.19\textwidth}
         \centering
         \includegraphics[width=\textwidth, height=2.5cm]{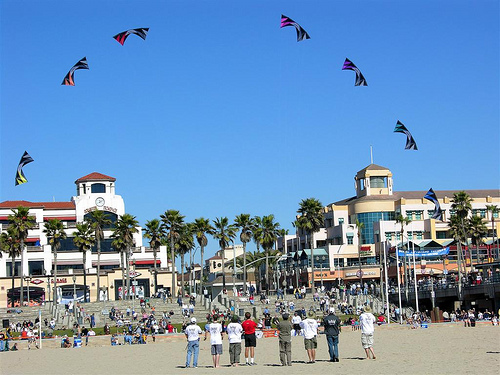}
         \caption{\small{\textbf{q}: Are those kites in the \textcolor{red}{air}/\textcolor{blue}{water}? \\ \textbf{a}: \textcolor{red}{yes}/\textcolor{blue}{no}.\vspace{\baselineskip}}}
         \label{137}
     \end{subfigure}
     \hfill
     \begin{subfigure}[b]{0.19\textwidth}
         \centering
         \includegraphics[width=\textwidth, height=2.5cm]{images/1792.png}
         \caption{\small{\textbf{q}: What is she wearing on her \textcolor{red}{head}/\textcolor{blue}{throat}? \\ \textbf{a}: \textcolor{red}{helmet}/\textcolor{blue}{nothing}.}}
         \label{1792}
     \end{subfigure}
     \hfill
     \begin{subfigure}[b]{0.19\textwidth}
         \centering
         \includegraphics[width=\textwidth, height=2.5cm]{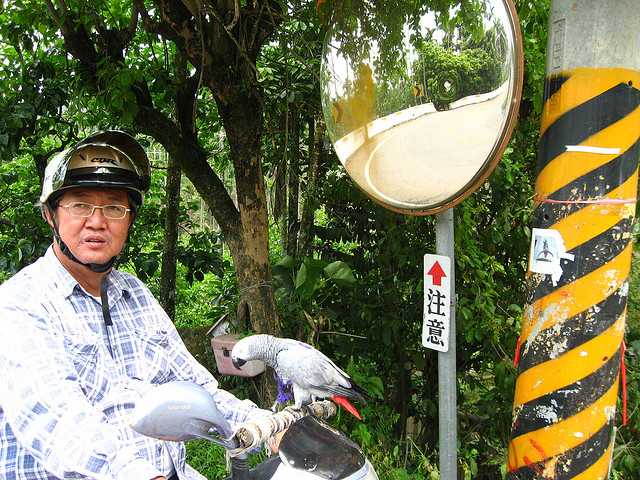}
         \caption{\small{\textbf{q}: What bird is on the man's \textcolor{red}{bike}/\textcolor{blue}{truck}? \\ \textbf{a}: \textcolor{red}{parrot}/\textcolor{blue}{parrot}.\vspace{\baselineskip}}}
         \label{800}
     \end{subfigure}
     \hfill
     \begin{subfigure}[b]{0.19\textwidth}
         \centering
         \includegraphics[width=\textwidth, height=2.5cm]{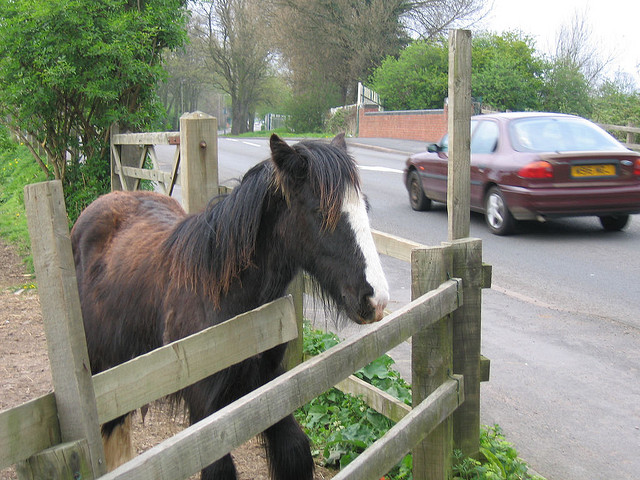}
         \caption{\small{\textbf{q}: What color is the \textcolor{red}{horse}/\textcolor{blue}{zebra}? \\ \textbf{a}: \textcolor{red}{brown}/\textcolor{blue}{black and white}.\vspace{\baselineskip}}}
         \label{1468}
     \end{subfigure}
        \caption{Local explanations for \textbf{Sibling Noun} counterfactual perturbations.}
        \label{Verbs photos}
\end{figure}

With reference to Sibling Noun substitutions, we notice as a global pattern that $M$ has insufficient separation ability when the sibling nouns refer to rooms of buildings or houses. As an example, in Figure \ref{2605}, $M$ cannot properly differentiate between the described interior spaces and can be easily fooled by substitutions involving places of different functionality. $M$ is also confused in the case of Figure \ref{800}, when sibling means of transport are substituted. Specifically, $M$ insists on its answer even though a concept not existing in the image appears (bike$\rightarrow$truck). This indicates that $M$ rather trespasses the linguistic modality context, providing an 'easy' answer based on the visual modality, since the only \textit{bird} appearing in the image is a \textit{parrot}. In this case, the relevant position of the bird \textit{on the man's bike/truck} is ignored.

An interesting behavior is observed when sibling concepts involving animals are tested, as in Figure \ref{1468}. In this case, $M$ seems to circumvent reasoning over the image, providing an answer based on knowledge it has most possibly acquired during its pre-training phase (zebras are black and white in color). Nevertheless, $M$ is not fooled by the horse$\rightarrow$zebra substitution, in which case it would conclude that \textit{the zebra is brown}, which is a wrong factual statement.
Another case that $M$ is not being fooled is depicted in Figure \ref{1792}. In this case, $M$ is very consistent in sibling entities that declare human body parts, which means that it correctly perceives their differences and does not group them in an arbitrary way. Furthermore, $M$ presents a correct reasoning process by differentiating its answer when the sibling concepts present very different meanings between them, as the concepts air$\rightarrow$water in Figure \ref{137}.

Overall, Siblings Noun substitution provided a rich set of insights, unequally relying on either $q$ or $I$ to derive an answer in many cases, rather than providing an uncertain outcome (such as answering "nothing" in the examples of Figures \ref{800}, \ref{1468}, similarly to the correct reasoning of Figure \ref{1792}). In total, this indicates an unstable behavior of $M$ towards different sibling pairs, yielding unpredictable outcomes under different substitutions of the same conceptual distance.

\subsection{Deletion Noun explanations}

\begin{figure}[ht!]
     \centering
     \begin{subfigure}[b]{0.22\textwidth}
         \centering
         \includegraphics[width=\textwidth, height=3cm]{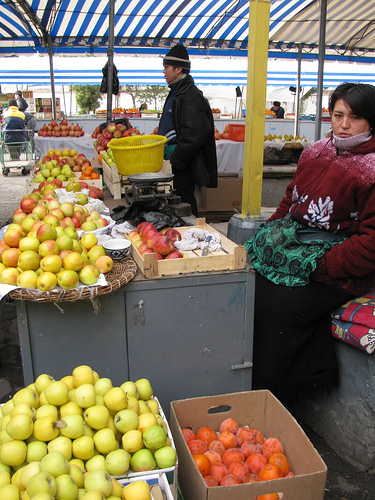}
         \caption{\small{\textbf{q}: Is the woman's \st{hair} tied back? \\ \textbf{a}: \textcolor{red}{no}/\textcolor{blue}{no}.\vspace{\baselineskip}}}
         \label{3891}
     \end{subfigure}
     \hfill
     \begin{subfigure}[b]{0.22\textwidth}
         \centering
         \includegraphics[width=\textwidth, height=3cm]{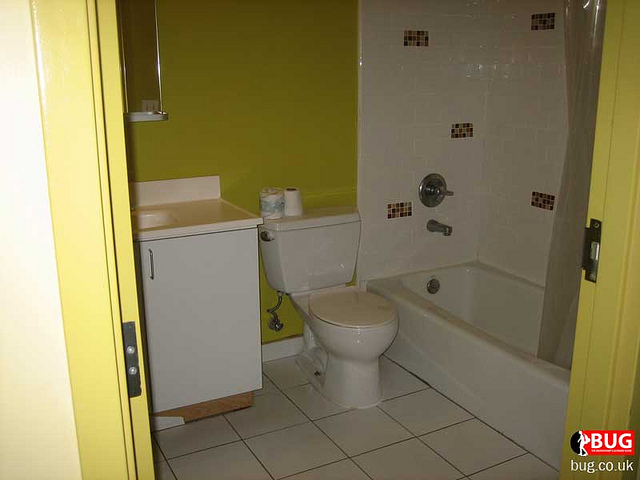}        
         \caption{\small{\textbf{q}: What color is the \st{bathroom}? \\ \textbf{a}: \textcolor{red}{yellow}/\textcolor{blue}{white}.\vspace{\baselineskip}}}
         \label{331}
     \end{subfigure}
     \hfill
     \begin{subfigure}[b]{0.22\textwidth}
         \centering
         \includegraphics[width=\textwidth, height=3cm]{images/2363.png}
         \caption{\small{\textbf{q}: Where are the \st{cakes}? \\ \textbf{a}: \textcolor{red}{table}/\textcolor{blue}{table}.\vspace{\baselineskip}}}
         \label{2363}
     \end{subfigure}
     \hfill
     \begin{subfigure}[b]{0.22\textwidth}
         \centering
         \includegraphics[width=\textwidth, height=3cm]{images/99.png}
         \caption{\small{\textbf{q}: How many animals are in this \st{photo}? \\ \textbf{a}: \textcolor{red}{2}/\textcolor{blue}{2}.}}
         \label{99}
     \end{subfigure}
        \caption{Local explanations for \textbf{Deletion Noun} counterfactual perturbations.}
        \label{Verbs photos}
\end{figure}

Regarding the counterfactual questions concerning deletions of nouns, we firstly observe the following pattern: When the deleted noun has a determining role in another noun already present in the question, $M$ maintains its original answer even after the deletion. Therefore, we detect a tendency towards attaching to the determined noun, while at the same time not paying due attention to the determiner noun. Hence, $M$ answers such questions arbitrarily, even though its answer cannot be perceived as wrong; a human could have also answered the same, especially in yes/no questions. A related example is provided in Figure \ref{3891}.

Another pattern that we detect concerns questions that refer to the color of a noun, which has been deleted. In these cases, $M$ tends to respond with the most dominant color in the image, without taking into account the absence of the noun that this color should define, as in Figure \ref{331}. Of course, we regard this behavior as justified, since a human would most probably answer such questions in the same way.
Similarly, in questions concerning the location of a noun, which has been deleted, $M$ answers with the most dominant entity present in the given image. A relevant example is demonstrated in Figure \ref{2363}.

Finally, we list some nouns to which we notice that $M$ does not pay due attention when asked to give an answer, as in Figure \ref{99}. Specifically, even after deleting them, $M$ tends to return the initial answer with great frequency. These words are: image, photographs, human, man, animal, room. In general, these are words that are encountered very often in questions and usually act in addition to other, more specific, entities. Once again, this behavior is justified.

All in all, we observe that the random deletion of a noun results in a rather expected model behavior, driven by dominant visual concepts present in the given image.

\section{Conclusion and Future Work}

Counterfactual perturbations in VQA models can provide novel and useful insights regarding model robustness and explainability of results. In our work, we
propose a knowledge-based counterfactual framework targeting substitutions on questions. Specifically, our framework suggests multiple types of word-level linguistic transformations in order to probe selected VQA models in a black-box fashion, and investigate whether the presence of counterfactual questions will lead to unexpected model responses. Through this process, underlying linguistic biases are revealed, while informative explanations regarding the model's behavior are provided, by extracting global rules in a qualitative manner, ultimately depicting those existing biases.
Our results on Visual Genome and VQA-v2 datasets, using ViLT model as proof of concept, illustrate the
merits for our approach, highlighting concepts that incite model biases, in a model-agnostic manner. As an immediate extension of our method, we aim to apply the same linguistic perturbations on dataset answers, addressing VQA models that reason over multiple choice answers. As future work, we plan to expand our approach to other related visiolinguistic tasks, such as Text - Image Retrieval, Visual
Entailment, and Visual Commonsense Reasoning, while another direction involves crafting counterfactual perturbations targeting the visual modality.

\begin{acknowledgments}
The research work was supported by the Hellenic Foundation for Research and Innovation (HFRI) under the 3rd Call for HFRI PhD Fellowships (Fellowship Number 5537).  
\end{acknowledgments}


\bibliography{sample-ceur}

\appendix

\end{document}